%% file: root.tex
\documentclass[letterpaper, 10 pt, conference]{ieeeconf}  %

\IEEEoverridecommandlockouts                              %

\usepackage{amsmath,amssymb,amsfonts}
\usepackage{graphicx}
\usepackage{subcaption}
\usepackage{booktabs}
\usepackage{multirow}
\usepackage{makecell}   %
\usepackage{cuted}      %
\usepackage{capt-of}
\usepackage{xcolor}
\usepackage[colorlinks=true,allcolors=blue]{hyperref}

\graphicspath{{figures/}%
              {sections/figure/}%
              {sections/01_introduction/}%
              {sections/02_related_work/}%
              {sections/03_problem_statement/}%
              {sections/04_method/}%
              {sections/05_experiment/}%
              {sections/06_discussion/}}

\newcommand{\std}[1]{\ensuremath{{\scriptstyle\pm#1}}}

\title{\LARGE \bf
PatternDex: Learning Interaction Patterns to Guide Reinforcement Learning of Bimanual Dexterous Manipulation of Articulated Objects}

\author{
David Minkwan Kim,
Runfa Blark Li,
Beckham Po-Ju Lee,
Nikolay Atanasov,
Truong Nguyen \\
University of California, 
San Diego
}

\begin{document}
\maketitle
\thispagestyle{empty}
\pagestyle{empty}

\input{sections/figure/teaser}

\input{sections/00_abstract/abstract}

\input{sections/01_introduction/introduction}
\input{sections/02_related_work/related_work}

\input{sections/03_problem_statement/problem_statement.tex}
\input{sections/04_method/method}

\input{sections/05_experiment/experiment}

\input{sections/06_conclusion/conclusion}

\addtolength{\textheight}{-12cm}   %

\bibliographystyle{IEEEtran}
\bibliography{IEEEabrv,references}

\end{document}

%% file: sections/figure/teaser.tex
\newcommand{\teaserwidth}{1.0}
\begin{strip}
  \centering
  \includegraphics[width=\teaserwidth\textwidth]{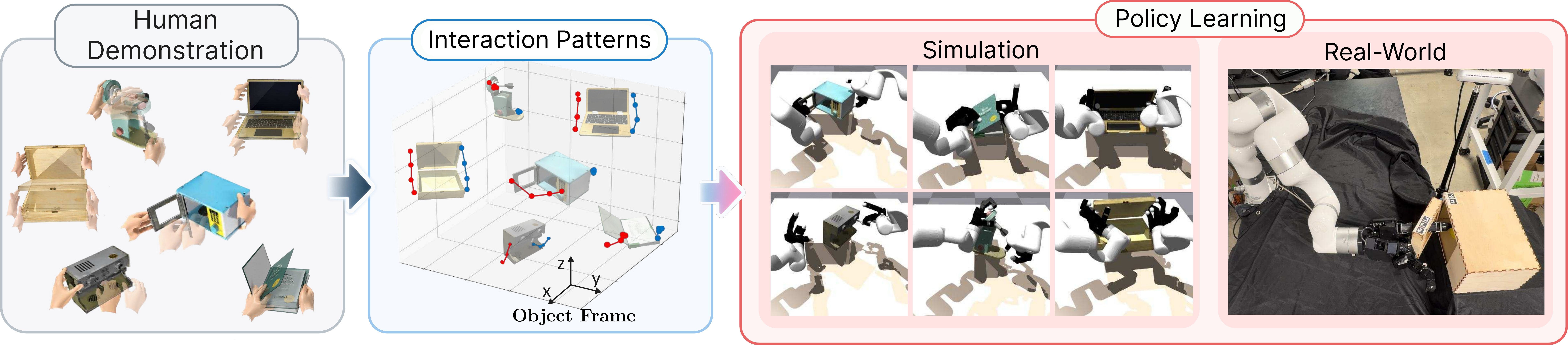}
  \captionof{figure}{\textbf{Overview of PatternDex.} Given a human demonstration, PatternDex learns the interaction pattern that describes the relationship between the object trajectory and the wrist motions in the object frame. Since this pattern is dictated by the object rather than the hands, it is agnostic to hand embodiments. With this pattern as guidance, PatternDex trains a policy that manipulates articulated objects with the target robot hand in both simulation and the real world. The target robot is a bimanual setup of Allegro hands mounted on xArm6 arms.}
  \label{fig:teaser}
\end{strip}

%% file: sections/00_abstract/abstract.tex
\begin{abstract}
In this paper, we develop a method that enables bimanual dexterous hands to manipulate articulated objects with a high success rate without suffering from an embodiment gap. We observe that the correlation between hand motions and object motions is dictated by the object rather than the hands and can be learned from human-object demonstrations. Based on this observation, we propose PatternDex, a method that learns this correlation and represents it as a token sequence, which we call an interaction pattern. From this pattern, PatternDex estimates the wrist motions and contact points that fit the target robot, and then trains a reinforcement learning policy that exploits these estimates as guidance. Since the guidance fits the target embodiment, the policy explores only the actions that the target robot can execute and thus achieves high success rates. PatternDex also requires only simple fine-tuning to train a new robot, since it can reuse the learned interaction pattern. We evaluate PatternDex with bimanual dexterous hands on human demonstrations from the ARCTIC dataset. PatternDex achieves, on average, a 92.8\% success rate with Allegro hands, while the state-of-the-art baseline achieves 52.2\%. Also, it achieves success rates above 70\% with three other robot hands after fine-tuning alone. Furthermore, we verify that the learned policy transfers well to a real-world task of opening a microwave. Videos and additional results are available at \url{https://patterndex.github.io/PatternDex/}.

\end{abstract}

%% file: sections/01_introduction/introduction.tex
\section{INTRODUCTION}
\label{sec:introduction}

Dexterous manipulation of articulated objects, such as opening a microwave door or a laptop lid, is a challenging problem because it requires coordinating and synchronizing high-degree-of-freedom (DoF) hands.
Successful approaches for these tasks adopt imitation learning, which trains a reinforcement learning policy to reproduce the motions observed in human demonstrations~\cite{dapg, opentelevision,regrind, maniptrans,dexmachina, toporetarget, objdex}. 
A straightforward method is to train a policy with robot execution data collected through teleoperation, where a human operator controls the robot hands directly~\cite{dapg, opentelevision}. However, collecting teleoperation data is expensive and time-consuming.
Another approach is to use kinematic retargeting that maps the human finger joints to the robot finger joints~\cite{regrind, maniptrans, dexmachina, toporetarget}, but this method becomes unreliable when the embodiment gap between the human and the robot hands is large.
A more recent approach~\cite{objdex} observes that the human wrist motion is less sensitive to the embodiment gap than the finger motion, so it imitates only the human wrist motion and learns the finger actions with reinforcement learning. 
However, this method addresses the embodiment gap with respect to the fingers but not with respect to the hand size and structure. When the robot wrist is placed at the human wrist pose, a larger robot hand exceeds the manipulation workspace, and a smaller hand fails to reach the object. Thus, imitating the human wrist alone is still insufficient to eliminate the embodiment gap.

We observe that in articulated manipulation tasks, the correlation between the hand and object motions is dictated by the desired object trajectory, rather than the hands. The key idea of our work is to learn this relationship and to represent it as a sequence of tokens in a latent space. We refer to such a sequence as an \emph{interaction pattern}. Since the interaction pattern depends on the object rather than the hands, it is agnostic to the hand embodiment and can be mapped not only to human hands but also to various robot hands. 
We develop a model that learns the interaction \textbf{Pattern} from human-object demonstrations and decodes it into training guidance for a reinforcement learning (RL) policy to achieve \textbf{Dex}terous manipulation of articulated objects. We refer to our method as \textbf{PatternDex}. By encoding an interaction pattern and combining it with a robot hand description, PatternDex estimates wrist motions and contact points that fit the target robot hand and uses them as guidance to drive the policy's actions.
Fig.~\ref{fig:teaser} gives an overview of PatternDex, where the interaction pattern is learned from human demonstrations and reproduced with bimanual dexterous robot hands in simulation and the real world. Our contributions are summarized as follows.

\begin{itemize} 
 \item We learn a latent representation of bimanual object manipulation in the form of a token sequence, which we call an interaction pattern. This representation captures the hand-object correlation dictated by the desired object trajectory and is agnostic to specific hand embodiments.

 \item We design a guidance predictor that combines an interaction pattern with a robot hand description to predict the wrist motions and contact points for the desired manipulation task. Since the interaction pattern transfers across robot hands, the predictor is trained once and needs only simple fine-tuning for a new robot hand.
 
 \item We train an RL policy with the wrist motions and contact points from the predictor as guidance. Since the guidance fits the target robot hand, the policy learns actions within the robot's capabilities rather than imitating human motions that are not feasible for the robot. As a result, our guided policy achieves higher success rates than policies trained directly from human hand motions.
\end{itemize}

%% file: sections/02_related_work/related_work.tex
\section{RELATED WORK}
\label{sec:related_work}

\textbf{Dexterous Manipulation with Imitation Learning.}
Reinforcement learning has solved dexterous manipulation tasks by designing a reward that specifies what to achieve in each task~\cite{inhand, dextrahg, humanoiddex} or by using a pretrained controller that moves the fingers so that the policy only gives high-level commands~\cite{dexteritygen}. However, such methods are hard to design manually for long-horizon tasks such as manipulating an articulated object. Many works therefore learn a policy by imitating human hand-object demonstrations, which show how the human hand moves and where it touches the object~\cite{maniptrans, dexmachina, regrind, toporetarget, objdex}. ManipTrans~\cite{maniptrans} and DexMachina~\cite{dexmachina} kinematically retarget the human hand motion by matching a fixed set of keypoints on the human hand with the corresponding keypoints on the robot hand. ReGRIND~\cite{regrind} and TopoRetarget~\cite{toporetarget} formulate retargeting as an interaction-mesh optimization that preserves the spatial relationship between the hand and the object. Since these methods map each human finger to a robot finger, the retargeted motion becomes infeasible when a robot hand such as the large four-fingered Allegro hand follows the motion of the smaller five-fingered human hand. ObjDex~\cite{objdex} avoids this finger mapping by using only the human wrist trajectory as a reference, but the robot wrist is still placed at the human wrist pose. Therefore, a robot hand of a different size either exceeds the manipulation workspace or fails to reach the object. 
In contrast, we solve this problem by estimating wrist motions that fit the robot embodiment instead of using the human wrists directly. We further add contact points estimated for the same embodiment, so the policy is guided on where the wrist should go and what the fingers should touch.

\textbf{Hand-Object Motion Correlation.}
A line of work in computer graphics generates human hand motion from object motion. OMOMO~\cite{omomo} takes object motion alone and predicts where the hands should be to achieve it, and then generates a body pose from the predicted hands. CAMS~\cite{cams} uses the object shape and a sequence of object poses to predict the hand pose for each frame in a space defined with respect to the object and the contact. Both works show that a consistent correlation exists between object motion and the hand motion that achieves it, and that this correlation can be learned from human motion data. However, these methods generate the motion of human hands and have not been applied to a robot hand of a different embodiment. Accordingly, we learn this correlation as an embodiment-independent token sequence and apply it to robot hands by decoding the wrist motions and contact points for a given target hand.

%% file: sections/03_problem_statement/problem_statement.tex
\section{PROBLEM STATEMENT} \label{sec:problem_statement}
Given human hand-object demonstrations together with a description of the target robot hand, we address the problem of learning a control policy for bimanual dexterous manipulation of articulated objects.
Each demonstration provides a sequence of object states and human hand states. The object state includes the pose, the articulation angle of the object, and the object shape (e.g., mesh or point cloud). The human hand states include the wrist poses and the shape of the human hands.

This problem poses two challenges. The human and robot hands differ in size and structure, so direct transfer of the demonstrated hand states to the robot is infeasible. Also, since manipulation of articulated objects is contact-rich with a large action space, reinforcement learning with a reward alone is unreliable.
Thus, we decompose the problem into two stages to address these challenges: the prediction stage (Sec.~\ref{subsec:problem_first}) and the guided policy learning stage (Sec.~\ref{subsec:problem_statment_policy_learning}). 

\subsection{Robot Wrist and Object Contact Prediction Stage}
\label{subsec:problem_first}

We aim to estimate wrist motions and contact points feasible for the robot hands, using the demonstrated object trajectories and the human hand motions. The prediction stage uses the following inputs:
\begin{itemize}
    \item $\mathcal{O}_t = (o^{\text{p}}_t, o^{\text{r}}_t, o^{\text{j}}_t, \mathcal{P}^o_t)$: the object state at frame $t$, including its position $o^{\text{p}}_t \in \mathbb{R}^{3}$, orientation $o^{\text{r}}_t \in \mathbb{R}^{6}$ (represented as a 6-D vector that concatenates the first two columns of the rotation matrix, which gives a continuous representation~\cite{rot6d}), articulation angle $o^{\text{j}}_t \in \mathbb{R}$, and surface point cloud $\mathcal{P}^o_t \in \mathbb{R}^{N \times 3}$;

    \item $W_t \in \mathbb{R}^{2 \times 9}$: the human wrist poses at frame $t$, described by the 3-D position and the 6-D orientation of each hand relative to the object frame;

    \item $\mathcal{R} \in \mathbb{R}^{K \times d_{\mathcal{R}}}$: the robot hand description (e.g., parsed from a unified robot description format (URDF)) consisting of $K = N_f + 1$ rows specifying $d_{\mathcal{R}}$ geometric attributes (number of joints and length and width of each link) for the $N_f$ fingers and the palm.
\end{itemize}
Then, the prediction stage estimates the desired robot wrist motions and object contact points:
\begin{itemize}
    \item $\hat{W}_t \in \mathbb{R}^{2 \times 9}$: the estimated robot wrist motions at frame $t$, described in the same form as $W_t$;

    \item $\hat{C}_t \in \mathbb{R}^{2N}$: the estimated contact points, giving the distance from each of the $N$ object surface points to the surface of each robot hand.
\end{itemize}
In summary, this stage aims to train a prediction model $G$ that estimates $\hat{W}_t$ and $\hat{C}_t$ over a horizon of $H+1$ frames,
\begin{equation}
     G\big(\mathcal{O}_{t:t+H},\, W_{t:t+H},\, \mathcal{R}\big) = (\hat{W}_{t:t+H},\, \hat{C}_{t:t+H}).
\end{equation}
In Sec.~\ref{subsec:prediction encoder and decoder}, we discuss the design and training of $G$.

\subsection{Guided Policy Learning Stage}
\label{subsec:problem_statment_policy_learning}

We formulate the bimanual dexterous manipulation problem as a Markov Decision Process (MDP) defined by a tuple $(\mathcal{S}, \mathcal{A}, p, r, \gamma, T)$ with elements defined as follows.
\begin{itemize}
    \item $\mathcal{S}$ is the state space consisting of states $s_t = (q_t, \hat{\mathcal{O}}_t)$, where $q_t$ is the proprioception of the two robot hands given by the joint angles of the fingers and wrists, and $\hat{\mathcal{O}}_t = (\hat{o}^{\text{p}}_t, \hat{o}^{\text{r}}_t, \hat{o}^{\text{j}}_t)$ is the manipulated object state (position, orientation, and articulation angle in the same form as $\mathcal{O}_t$).

    \item $\mathcal{A}$ is the action space consisting of actions $a_t = (a^{\text{f}}_t, \Delta a^{\text{w}}_t)$, where $a^{\text{f}}_t$ is the finger action (target joint angles for the fingers of both hands) and $\Delta a^{\text{w}}_t$ is a residual wrist correction added to the estimated wrist pose $\hat{W}_t$.

    \item $p(s_{t+1} \mid s_t, a_t)$ is the transition probability determined by the contact dynamics of the robot hands and the articulated object, which is computed by the Isaac Gym simulator \cite{isaacgym} in practice.
    
    \item $r = r_{\text{obj}} + r_{\text{cont}}$ is the reward function, consisting of the object reward $r_{\text{obj}}(s_t, \mathcal{O}_t)$, measuring how closely the current object state $\hat{\mathcal{O}}_t$ tracks the reference $\mathcal{O}_t$, and the contact reward $r_{\text{cont}}(s_t, \hat{C}_t)$, measuring how well the robot hands touch the estimated contact points.

    \item $\gamma$ and $T$ are the discount factor and the time horizon, respectively.
\end{itemize}

At each timestep $t$, the policy receives the state $s_t$, the reference object trajectory $\mathcal{O}_{t:t+H}$, and the contact points $\hat{C}_{t:t+H}$ obtained in the prediction stage and provides an action as follows:
\begin{equation}
a_t \sim \pi_{\theta}(\cdot \mid s_t, \mathcal{O}_{t:t+H}, \hat{C}_{t:t+H}),
\label{eq:policy}
\end{equation}
where the finger command given to the robot is $a^{\text{f}}_t$ and the wrist command is $\hat{W}_t + \Delta a^{\text{w}}_t$.
The policy $\pi_{\theta}$ is trained to maximize the expected discounted return,
\begin{equation}
J(\theta) = \mathbb{E}_{\mathcal{O},\, \pi_{\theta}}
\!\!\left[\sum_{t=0}^{T-1} \gamma^{t}
\big( r_{\text{obj}}(s_t, \mathcal{O}_t)
    + r_{\text{cont}}(s_t, \hat{C}_t) \big)\!\right].
\label{eq:objective}
\end{equation}
In Sec.~\ref{subsec:policy}, we discuss the training method in detail.

%% file: sections/04_method/method.tex
\input{sections/figure/method_overall}

\section{PATTERNDEX}
\label{sec:method}

In this section, we introduce our model, PatternDex, to realize the two stages introduced in Sec.~\ref{sec:problem_statement}.
In the prediction stage (Sec.~\ref{subsec:prediction encoder and decoder}), PatternDex learns a relationship between the object trajectory and human wrist poses and then estimates the wrist motions and the contact points from this relationship. In the policy learning stage (Sec.~\ref{subsec:policy}), a policy guided by the estimates is trained for bimanual dexterous manipulation of articulated objects. 
Fig.~\ref{fig:method_framework} illustrates the overview of the PatternDex model.

\subsection{Prediction Stage: Encoder-Decoder Model}
\label{subsec:prediction encoder and decoder}

The key idea of our work is to learn a relationship between a desired object trajectory $\mathcal{O}_{1:T}$ and the wrist motions $W_{1:T}$ that achieve it. We formalize this relationship as a latent space representation in the form of a token sequence that encodes the mapping from $\mathcal{O}_{1:T}$ to $W_{1:T}$. We call this token sequence an \emph{interaction pattern} $P_I$.
Each token of $P_I$ encodes relative translation and rotation of $W_t$ with respect to $\mathcal{O}_t$ at frame $t$. The token sequence captures the required wrist pose changes as the object state progresses across frames. 
To learn the interaction pattern $P_I$, we design an encoder model $\mathcal{E}$ consisting of a spatial network $f$ and a temporal network $\mathcal{T}$.

\textbf{Interaction Pattern Encoder.}
The spatial network $f$ is a multilayer perceptron (MLP) that takes the object state $\mathcal{O}_t$ and the human wrist pose $W_t$ as inputs and outputs a spatial token $\tilde{\mathbf{x}}_t \in \mathbb{R}^{d}$ ($d = 512$) that encodes the relative pose of the human wrists with respect to the object. The object position $o^{\text{p}}_t$, orientation $o^{\text{r}}_t$, articulation angle $o^{\text{j}}_t$, and $W_t$ are jointly encoded into a $d$-dimensional vector, with layer normalization $\mathrm{LN}$ and sinusoidal positional encoding $E_{\text{pos}}(t)$ to produce the spatial token $\tilde{\mathbf{x}}_t$:
\begin{equation} \label{eq:encoder_tokenization}
    \tilde{\mathbf{x}}_t = \mathrm{LN}\Big( f\big(\mathcal{O}_t, W_t\big) \Big) + E_{\text{pos}}(t) \in \mathbb{R}^{d}.
\end{equation}
To capture how the spatial tokens influence one another over time, we stack them in a time sequence:
\begin{equation}
X = [\tilde{\mathbf{x}}_t, \dots, \tilde{\mathbf{x}}_{t+H}]^\top \in \mathbb{R}^{(H+1) \times d}
\end{equation}
and input them to the temporal network $\mathcal{T}$.

The temporal network $\mathcal{T}$ consists of $L_{\mathcal{T}} = 4$ identical layers, each with $N_h = 4$ self-attention heads followed by a single feed-forward network. The spatial tokens $X$ are fed into $\mathcal{T}$ as the initial hidden state $X^{(0)} = X$. From the first layer $\ell = 1$ to the last layer $\ell = L_{\mathcal{T}}$, the temporal network $\mathcal{T}$ learns the motion context, which is the mutual influence among the spatial tokens. Each layer $\ell$ takes the previous hidden state $X^{(\ell-1)}$ and computes the attention logits among its tokens. At the $h$-th self-attention head ($h = 1, \dots, N_h$) of layer $\ell$, the attention logits $A^{(\ell,h)}_{\mathcal{T}} \in \mathbb{R}^{(H+1) \times (H+1)}$ are calculated as
\begin{equation}
    A^{(\ell,h)}_{\mathcal{T}} = \frac{X^{(\ell-1)} W^{(\ell,h)}_{q}\,(X^{(\ell-1)} W^{(\ell,h)}_{k})^{\top}}{\sqrt{d_h}},
    \label{eq:encoder_self}
\end{equation}
where $W^{(\ell,h)}_{q}, W^{(\ell,h)}_{k} \in \mathbb{R}^{d \times d_h}$ are learnable query and key projections, respectively, and $d_h$ is the dimension of each attention head. 
The $(i,j)$-th element of $A^{(\ell,h)}_{\mathcal{T}}$ measures how much frame $j$ influences the representation of frame $i$. Thus, $\mathcal{T}$ learns the changes of the object state and the wrist poses between consecutive frames. The attention weights and outputs at each layer are computed in the same way as in a standard Transformer method \cite{transformer}. The final layer output $X^{(L_{\mathcal{T}})}$ becomes the interaction pattern:
\begin{equation}
    P_I = X^{(L_{\mathcal{T}})} \in \mathbb{R}^{(H+1) \times d}.
\label{eq:context_memory}
\end{equation}
The interaction pattern $P_I$ represents the relative pose of the wrists with respect to the object at each frame and its progression over the consecutive frames $t$ to $t+H$. Notably, $P_I$ contains information independent of human hands, so it can be shared across embodiments.

\textbf{Interaction Pattern Decoder.}  
The decoder $\mathcal{D}$ takes the robot hand description $\mathcal{R}$ and the interaction pattern $P_I$ to estimate the wrist motions and contact points that fit the target robot.
To concatenate $\mathcal{R}$ with $P_I$, we devise an embodiment encoder $\phi$ that takes $\mathcal{R}$ and produces embodiment tokens $T_E$ summarizing the geometric features of the robot hand:
\begin{equation}
    T_E = \phi(\mathcal{R}) \in \mathbb{R}^{K \times d},
\end{equation}
and we concatenate $T_E$ with $P_I$ into a single memory $M$:
\begin{equation}
    M = \big[\, P_I \,;\, T_E \,\big] \in \mathbb{R}^{(H+1+K) \times d}.
    \label{eq:decoder_memory}
\end{equation}

We follow the chunk-wise prediction of ACT~\cite{act}, which estimates $H+1$ frames of wrist motions at once ($H = 63$). A learnable query token $\mathbf{q}_t \in \mathbb{R}^{d}$ is assigned to each frame, and we stack them in a time sequence into the query chunk $Q = [\mathbf{q}_1 ; \dots ; \mathbf{q}_{H+1}] \in \mathbb{R}^{(H+1) \times d}$, which enters the decoder as the initial hidden state $Z^{(0)} = Q + E_{\text{pos}}$.
The decoder $\mathcal{D}$ consists of $L_{\mathcal{D}} = 6$ identical layers, each with $N_h$ heads. Each layer $\ell$ applies self-attention to the previous hidden state $Z^{(\ell-1)}$ to obtain $\tilde{A}^{(\ell)}$, then cross-attention between $\tilde{A}^{(\ell)}$ and the memory $M$, and finally a feed-forward network to produce the hidden state $Z^{(\ell)}$. At the $h$-th cross-attention head of layer $\ell$, the cross-attention logits $A^{(\ell,h)}_{\text{cross}} \in \mathbb{R}^{(H+1) \times (H+1+K)}$ are calculated as
\begin{equation}
    A^{(\ell,h)}_{\text{cross}} = \frac{\tilde{A}^{(\ell)} \tilde{W}^{(\ell,h)}_{q}\,(M \tilde{W}^{(\ell,h)}_{k})^{\top}}{\sqrt{d_h}},
    \label{eq:decoder_cross2}
\end{equation}
where $\tilde{W}^{(\ell,h)}_{q}, \tilde{W}^{(\ell,h)}_{k} \in \mathbb{R}^{d \times d_h}$ are the learnable query and key projections of head $h$, and $d_h$ is the head dimension. 
The cross-attention logits $A^{(\ell,h)}_{\text{cross}}$ capture how much each token in $M$ contributes to the estimate of the robot wrist motions and contact points at each frame. 
Lastly, the final hidden state $Z^{(L_{\mathcal{D}})}$ is fed into the wrist head and the contact head, which decode the robot wrist motions $\hat{W}$ and the contact points $\hat{C}$, respectively.

Since $P_I$ is agnostic to embodiment, estimating the wrist motions and contact points for a new robot hand requires only fine-tuning $\mathcal{D}$ and $\phi$, without training $\mathcal{E}$ again.
Fig.~\ref{fig:pattern_fig} shows an example of an interaction pattern learned from a microwave-opening trajectory, together with the wrist motions estimated for Allegro.

\input{sections/figure/pattern_fig}

\textbf{Training.} To train the predictor $G$, we collect ground-truth robot wrist motions and contact points for various robot hand embodiments and tasks. We retarget human demonstrations from the ARCTIC dataset \cite{arctic} to the robot hands in the same way as in \cite{regrind}. From the retargeted result, we take only the wrist motions $W^{R}_t$ and the contact points $C^{R}_t$, leaving out the finger motions. Even if the retargeted results are imprecise, the policy in Sec.~\ref{subsec:policy} corrects the estimated wrist motions and contact points during training. The estimated $\hat{W}_t$ and $\hat{C}_t$ are supervised by $W^{R}_t$ and $C^{R}_t$, and the loss functions for both are mean-squared errors over the chunk of $H+1$ frames:
\begin{gather}
\mathcal{L}_{\text{wrist}} = \frac{1}{H+1} \sum_{\tau=t}^{t+H} \| \hat{W}_\tau - W^{R}_\tau \|_2^2, \notag\\[2pt]
\mathcal{L}_{\text{cont}} = \frac{1}{H+1} \sum_{\tau=t}^{t+H} \| \hat{C}_\tau - C^{R}_\tau \|_2^2,
\label{eq:prior_losses}
\end{gather}
and the total loss is $\mathcal{L}_{\text{total}} = \mathcal{L}_{\text{wrist}} + \mathcal{L}_{\text{cont}}$.
With the wrist motions $\hat{W}$ and the contact points $\hat{C}$ estimated by $G$, we next describe how they guide the policy learning.

\subsection{Guided Policy Learning Stage: Reinforcement Learning with Guidance}
\label{subsec:policy}

We train a bimanual dexterous policy using guidance from $\hat{W}$ and $\hat{C}$ obtained in the prediction stage.
While the reference object trajectory $\mathcal{O}$ specifies only how the object should move, the guidance from $\hat{W}$ and $\hat{C}$ additionally specifies where the wrist should be and what the fingers should touch, which makes the policy training much more efficient.
We use the PPO algorithm \cite{ppo} with the object trajectory $\mathcal{O}_{t:t+H}$, the contact points $\hat{C}_{t:t+H}$, and the current state $s_t$ as inputs to the policy at each time $t$. The policy outputs an action $a_t = (a^{\text{f}}_t, \Delta a^{\text{w}}_t)$ that drives the finger joints of both hands ($a^{\text{f}}_t$) and provides a residual wrist correction ($\Delta a^{\text{w}}_t$) to $\hat{W}_t$. The goal is to drive the manipulated object state $\hat{\mathcal{O}}_t$ to the reference object state $\mathcal{O}_t$ at every frame by maximizing two rewards: an object reward $r_{\text{obj}}$ and a contact reward $r_{\text{cont}}$.

\textbf{Object Reward.}
The object reward $r_{\text{obj}}$ measures how closely the manipulated object state $\hat{\mathcal{O}}_t$ tracks the reference state $\mathcal{O}_t$ in translation, rotation, and articulation angle:
\begin{gather}
r_{\text{obj}} = \exp\!\big( -\lambda_{\text{p}} \, r_{\text{p}} - \lambda_{\text{r}} \, r_{\text{r}} - \lambda_{\text{j}} \, r_{\text{j}} \big), \notag\\[2pt]
r_{\text{p}} = \lVert \hat{o}^{\text{p}}_t - o^{\text{p}}_t \rVert_{2}, \;\;
r_{\text{r}} = d_{\text{r}}(\hat{o}^{\text{r}}_t, o^{\text{r}}_t), \;\;
r_{\text{j}} = \lvert \hat{o}^{\text{j}}_t - o^{\text{j}}_t \rvert, \label{eq:object_reward}
\end{gather}
where $\hat{o}^{\text{p}}_t$, $\hat{o}^{\text{r}}_t$, $\hat{o}^{\text{j}}_t$ denote the position, rotation, and articulation angle of $\hat{\mathcal{O}}_t$, respectively.
The rotation distance is $d_{\text{r}}(\hat{o}^{\text{r}}_t, o^{\text{r}}_t) = 2\cos^{-1}\!\big(\lvert \hat{\text{q}}_t^{\top} \text{q}_t \rvert\big)$, where $\hat{\text{q}}_t, \text{q}_t \in \mathbb{R}^4$ are the unit quaternions of $\hat{o}^{\text{r}}_t$ and $o^{\text{r}}_t$, and $\hat{\text{q}}_t^{\top} \text{q}_t$ is their Euclidean inner product~\cite{rotmetrics}.
Since the joint angle of every articulated object is limited to $[0, \pi]$, the absolute difference $r_{\text{j}}$ is the angular distance.
The weights $\lambda_{\text{p}}$, $\lambda_{\text{r}}$, and $\lambda_{\text{j}}$ balance the different units of the three terms (m and rad), and we use $\lambda_{\text{p}} = 20$, $\lambda_{\text{r}} = 2$, and $\lambda_{\text{j}} = 2$.

\textbf{Contact Reward.}
The contact reward $r_{\text{cont}}$ measures how well each robot hand $h \in \{\mathrm{L},\mathrm{R}\}$ touches the estimated contact points $\hat{C}$. 
To reduce the computation, we collect the points from $\hat{C}$ whose contact distance is below $\epsilon = 2$ cm into the target set $\hat{C}_h = \{p_1, \dots, p_{K_h}\}$ for each hand $h$.
On the robot hand, we use two sets of keypoints, the fingertip set $\mathcal{F}_h \subset \mathbb{R}^3$ and the set $\mathcal{P}_h \subset \mathbb{R}^3$ that includes $\mathcal{F}_h$ and the palm center.
The contact reward $r_{\text{cont}}$ consists of a coverage term $r_{\text{cov}}$, which increases as the keypoints approach the target points, and a penalty term $r_{\text{pen}}$, which penalizes the fingertips that touch the object points far from the target points. The coverage term is:
\begin{equation}
r_{\text{cov}} = \sum_{h \in \{\mathrm{L},\mathrm{R}\}} \frac{1}{K_h} \sum_{p \in \hat{C}_h} \exp\!\Big( -\frac{1}{\sigma} \min_{q \in \mathcal{P}_h} \lVert p - q \rVert_2 \Big),
\label{eq:coverage_reward}
\end{equation}
where $\sigma = 3$ cm. The penalty term is
\begin{align}
&r_{\text{pen}} = -\sum_{h \in \{\mathrm{L},\mathrm{R}\}} \min\Big\{ \sum_{j \in \mathcal{F}_h} c_j \, \delta_j,\ 3 \Big\}, \label{eq:penalty_reward}\\[4pt]
&\hspace{0.2cm} \text{where }\delta_j =
\begin{cases}
1, & \text{if } \min_{p \in \hat{C}_h} \lVert q_j - p \rVert_2 > d_{\text{pen}}=10\;\text{cm} \\
0, & \text{otherwise}
\end{cases}\notag
\end{align}
and $c_j \in \{0,1\}$ indicates whether fingertip $q_j \in \mathcal{F}_h$ contacts the object. We use $c_j$ to distinguish whether a fingertip touches a wrong part of the object or is merely approaching the target points.
The penalty magnitude is clamped to $3$ so that it does not overwhelm $r_{\text{cov}}$. The two terms are combined as
\begin{equation}
r_{\text{cont}} = w_{\text{cov}} \, r_{\text{cov}} + w_{\text{pen}} \, r_{\text{pen}},
\label{eq:contact_reward}
\end{equation}
with weights $w_{\text{cov}} = 0.25$ and $w_{\text{pen}} = 0.1$.
The policy is trained to maximize the total reward $r = r_{\text{obj}} + r_{\text{cont}}$.

Since the policy takes $\hat{C}_{t:t+H}$ as an input that guides the finger action $a^{\text{f}}_t$ toward the target contact points and adds the residual wrist correction $\Delta a^{\text{w}}_t$ to $\hat{W}_{t}$, the policy learns actions that manipulate the object within the workspace of the target robot hand. As a result, the policy achieves a high success rate, as we demonstrate in Sec.~\ref{subsec:eval_policy}.

%% file: sections/figure/method_overall.tex
\begin{figure*}[t]
  \centering
  \includegraphics[width=1\textwidth]{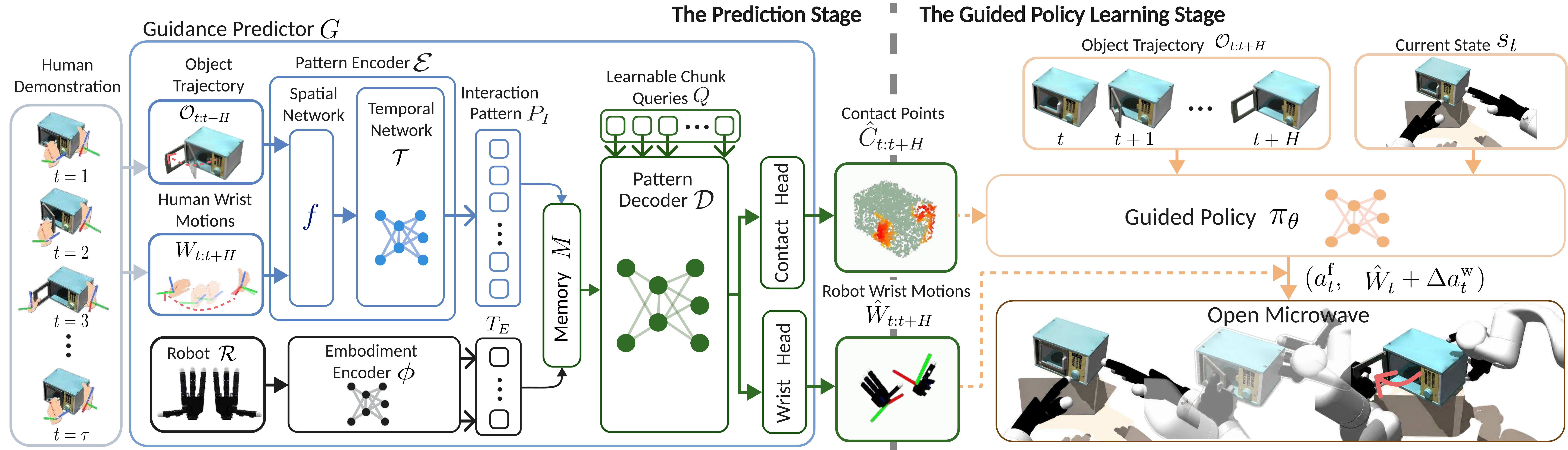}
  \caption{\textbf{PatternDex Model.} In the prediction stage (left), the guidance predictor $G$ encodes the object trajectory $\mathcal{O}_{t:t+H}$ and the human wrist motions $W_{t:t+H}$ into the interaction pattern $P_I$. From $P_I$ and the robot hand description $\mathcal{R}$, it then estimates the wrist motions $\hat{W}_{t:t+H}$ and the contact points $\hat{C}_{t:t+H}$. In the guided policy learning stage (right), the policy trained with these estimates executes bimanual dexterous manipulation of articulated objects, such as opening a microwave with Allegro hands mounted on xArm6 arms.}
  \label{fig:method_framework}
  \vspace{-8pt}
\end{figure*}

%% file: sections/figure/pattern_fig.tex
\begin{figure}[t]
  \centering
  \includegraphics[width=1\columnwidth]{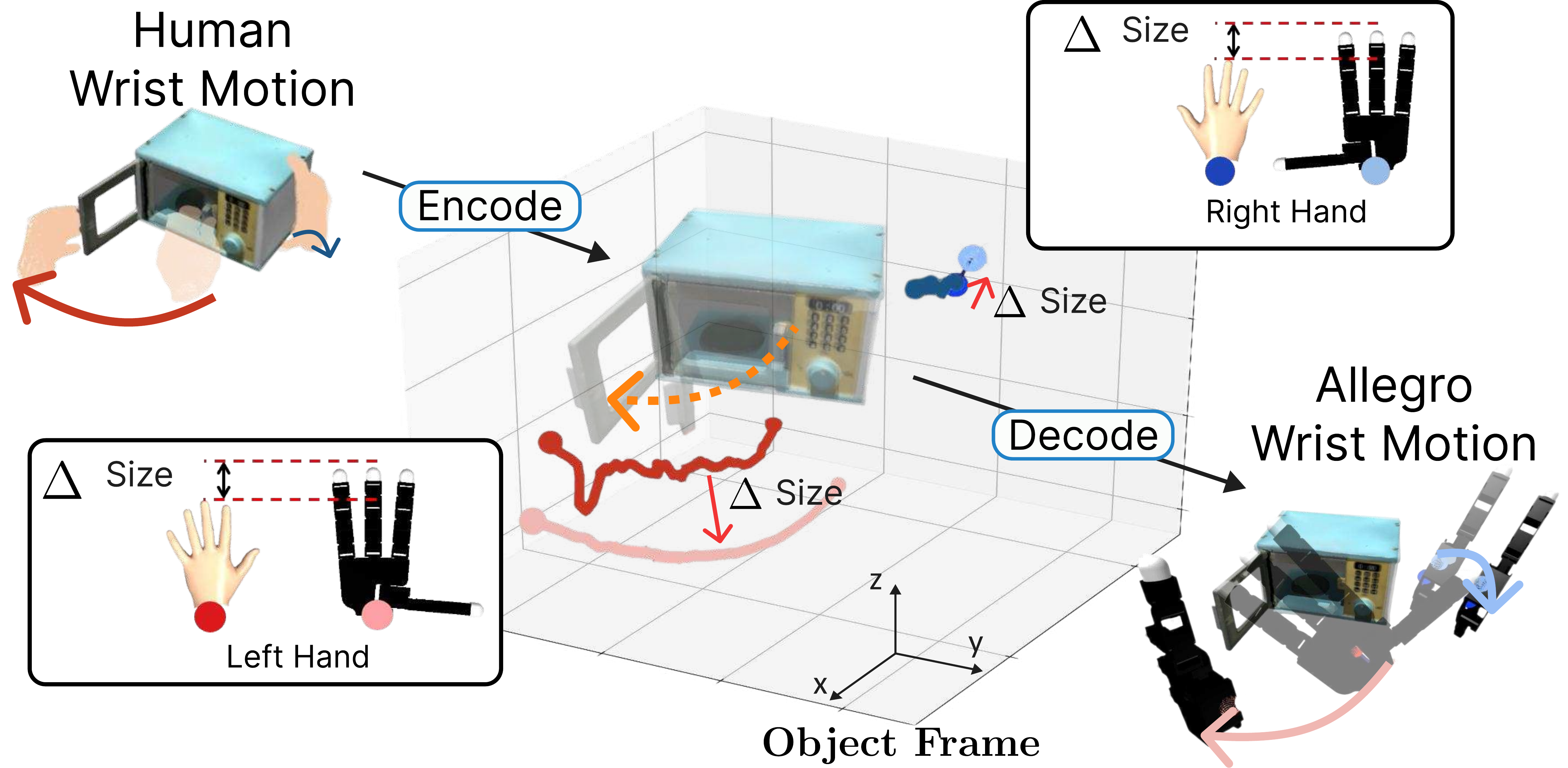}
  \caption{\textbf{Interaction pattern for opening the microwave.} The object trajectory is shown with its articulation angle only (dashed orange). The human wrist motions (red for the left hand, blue for the right hand) change as the articulation angle increases, and the interaction pattern encodes this change. The light red and light blue curves are Allegro wrist motions estimated from the interaction pattern. They preserve the shape of the human wrist motions and are shifted back according to the size of Allegro, which shows that the estimated wrist motions are adjusted to the robot.}
  \label{fig:pattern_fig}
  \vspace{-8pt}
\end{figure}

%% file: sections/05_experiment/experiment.tex
\section{EVALUATION}
\label{sec:experiment}
This section evaluates PatternDex on how accurately the guidance predictor $G$ estimates the wrist motions $\hat{W}$ and the contact points $\hat{C}$ that fit the target robot hand, and how successfully the guided policy reproduces the demonstrated manipulation in both simulation and the real world. PatternDex is trained on 229 human hand-object demonstrations from the ARCTIC dataset~\cite{arctic} and evaluated on 6 unseen demonstrations. All policies are trained in the Isaac Gym simulator~\cite{isaacgym}.

\subsection{Wrist Motion and Contact Point Prediction}\label{subsec:eval_prediction}

Table~\ref{tab:prior_human_eval} reports the interaction pattern consistency, the wrist offset, and the contact precision for four structurally different robot hands.
The pattern consistency is calculated as the Pearson correlation~\cite{pearson} between the frame-to-frame displacements of the human wrist and the predicted robot wrist along each of the $x$, $y$, $z$ axes in the object frame. Shadow, which is structurally closest to the human hand, shows the highest consistency of 99.1\%, and even Allegro, which differs the most, reaches 96.5\%. All four hands thus exceed 96\%.

The wrist offset is the distance between the human wrist and the estimated robot wrist. Allegro and Paxini are placed about 5 cm behind the human wrist, while Shadow and XHand are placed about 2.8 cm behind. These offsets reflect the structure of each hand rather than its overall size. Shadow and XHand have both the palm and the fingers about 1--2 cm longer than the human hand, and their wrists are placed about 2.8 cm behind, which roughly matches this uniform increase in length. Allegro has fingers about 6 cm longer than the human fingers but a palm of similar length, while Paxini has a palm about 5 cm longer than the human palm but fingers only about 2 cm longer. Their wrists are placed about 5 cm behind, which matches the more elongated part of each hand rather than the sum of the two.

The contact precision is the overlap between the estimated contact points and the human contact points, and all four hands reach 74--80\%. These results show that the predictor $G$ maintains the demonstrated interaction pattern while adapting the wrist motions and contact points to each robot embodiment.

\subsection{Guided Policy Evaluation}
\label{subsec:eval_policy}

We compare three types of guidance on six demonstrations with two Allegro hands mounted on xArm6 arms. Each episode starts with both hands about 22 cm away from the object. The retargeting methods~\cite{regrind, maniptrans, dexmachina, toporetarget} instead reset the hand to a state on the retargeted reference within the workspace, so the wrist placement is fixed by the reset rather than by learning. We therefore compare with ObjDex~\cite{objdex}, which starts from the same poses and guides the wrist with the human wrist motions. PatternDex guides the policy with the estimated wrist motions alone or with both the wrist motions and the contact points.
Each episode follows a 500-step reference object trajectory and counts as a success if the object stays within 10 cm in position, $30^\circ$ in rotation, and $10^\circ$ in joint angle of the reference without collision between the hands and arms. Table~\ref{tab:policy_eval} reports the success rate and the errors averaged over the steps before each episode ends.
ObjDex achieves a mean success rate of 52.2\% across the six demonstrations. With the estimated wrist motions alone, PatternDex reaches 58.1\%, 5.9\% higher than ObjDex. PatternDex guided by both the wrist motions and the contact points achieves the highest success rate of 92.8\%.

\input{sections/05_experiment/table1}
\input{sections/figure/policy_result}
\input{sections/05_experiment/table2}

The Allegro hand is larger than the human hand, so its fingers exceed the workspace for manipulating the object when the wrist is placed at the human wrist pose. Thus, ObjDex often explores actions that are not feasible for Allegro, which leads to its low success rate.
In contrast, PatternDex drives the policy's actions to the proper workspace, since it places the wrist according to the structure of the Allegro hand. The contact points further guide the policy on where to touch, which makes the training more efficient and raises the success rate.

\subsection{Cross-Embodiment Evaluation}\label{subsec:eval_embodiment}
\input{sections/05_experiment/table3}
Table~\ref{tab:embodiment_eval} shows the cross-embodiment success rates of the interaction pattern transferred from the Allegro hand to the Shadow, Paxini, and XHand hands.
To evaluate whether the interaction pattern learned with one embodiment can be reused for others, we train the entire PatternDex with the Allegro hand, and then freeze the pattern encoder $\mathcal{E}$ and fine-tune only $\mathcal{D}$ and $\phi$ for each of the other three hands.

Allegro trained with the entire method achieves a highest average success rate of 92.8\%, and Shadow, Paxini, and XHand transferred with the frozen $\mathcal{E}$ still exceed 70\%, reaching 84.0\%, 73.5\%, and 71.6\%, respectively.
Fig.~\ref{fig:policy_result} (a) shows the Allegro hand opening the microwave, the laptop, and the notebook through the full pipeline of PatternDex.
Fig.~\ref{fig:policy_result} (b) shows that the other three hands succeed on the same tasks without retraining $\mathcal{E}$.
\input{sections/figure/real_policy}

\subsection{Sim-to-Real Experiment}
\label{sim_to_real}
We deploy PatternDex on a real robot to verify that the policy trained in simulation transfers to the real world.
Since PatternDex has already learned the interaction pattern from the ARCTIC dataset, it needs only a single human demonstration video to let the target robot imitate a new task without collecting robot demonstrations.
Our example task is opening a real microwave with an Allegro hand mounted on an xArm6 arm.

From the video demonstration of a human opening the microwave, CoTracker3~\cite{cotracker3} tracks the position and the hinge angle of the microwave, and HaMeR~\cite{hamer} tracks the human wrist pose.
The tracked trajectories are fed into the guidance predictor $G$, and the guided policy is then trained in the Isaac Gym simulator with the estimated wrist motions and contact points. Fig.~\ref{fig:real_policy} (a) shows the overall pipeline of the sim-to-real experiment.
At each step, the policy receives the hinge angle of the microwave, the measured joint states of the arm and hand, and their velocities estimated by the PhysX engine of Isaac Gym, and outputs an action applied to Allegro and xArm6. Fig.~\ref{fig:real_policy} (b) shows snapshots of the real Allegro hand opening the microwave. The door angle increases from $0^\circ$ to $74.2^\circ$ over the 100 steps, and the robot opens the door in 7 of 10 trials, which verifies that the policy trained in the simulator transfers to the real robot.

%% file: sections/05_experiment/table1.tex
\begin{table}[t]
\centering
\caption{\textbf{Quantitative results of the predicted wrist poses and contact points on unseen demonstrations.} $\Delta$Palm and $\Delta$Finger are the differences in palm and finger lengths between each robot hand and the human hand (MANO~\cite{mano}).}
\label{tab:prior_human_eval}
\footnotesize\setlength{\tabcolsep}{3.5pt}
\begin{tabular}{l c c c c c}
\toprule
 & $\Delta$Palm & $\Delta$Finger & Pattern & Wrist & Contact \\
Hand & (cm) & (cm) & consist. (\%) $\uparrow$ & offset (cm) & prec. (\%) $\uparrow$ \\
\midrule
Allegro & +0.7 & +5.8 & 96.5\std{2.0} & 5.06\std{0.72} & 73.8 \\
Shadow  & +1.1 & +2.1 & 99.1\std{0.7} & 2.82\std{0.58} & 80.1 \\
Paxini  & +4.6 & +2.4 & 97.7\std{1.7} & 5.22\std{0.57} & 75.8 \\
XHand   & +1.9 & +2.3 & 97.7\std{4.5} & 2.77\std{0.39} & 77.4 \\
\bottomrule
\end{tabular}
\vspace{-8pt}
\end{table}

%% file: sections/figure/policy_result.tex
\begin{figure*}[t]
  \centering
  \includegraphics[width=\textwidth]{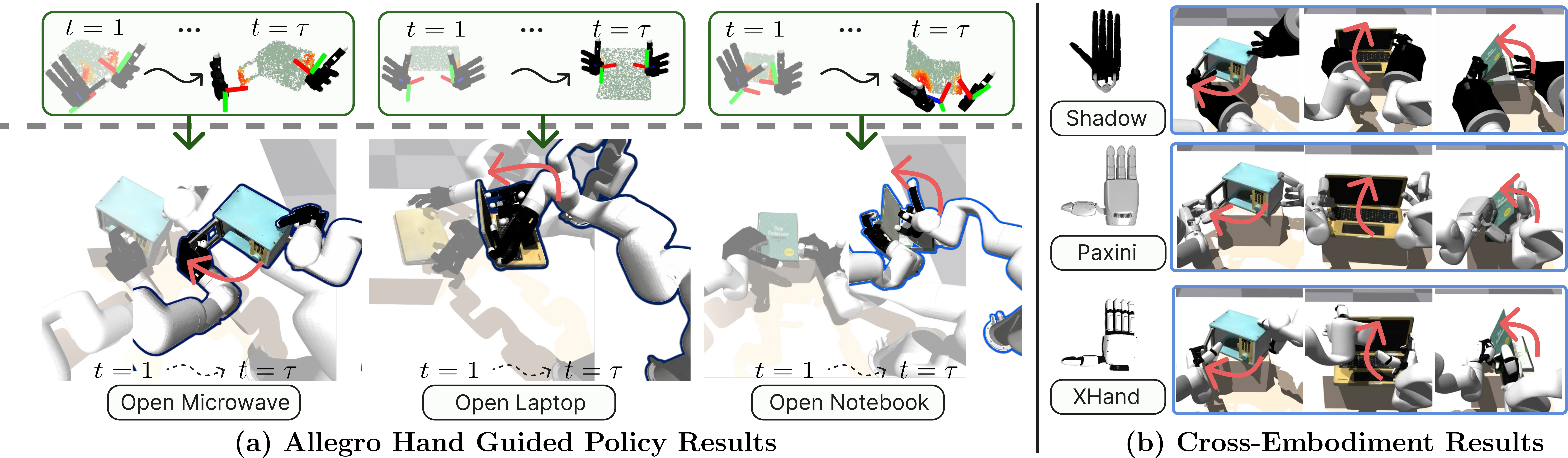}
  \caption{\textbf{Qualitative results of the guided policy.} (a) Guided by the estimated wrist motions and contact points shown in the top row, Allegro hands open the microwave, the laptop, and the notebook. (b) Shadow, Paxini, and XHand hands succeed on the same tasks with the interaction pattern transferred from the Allegro hand.}
  \label{fig:policy_result}
\end{figure*}

%% file: sections/05_experiment/table2.tex
\begin{table*}[t]
\centering
\caption{\textbf{Quantitative results of the guided policy for bimanual dexterous manipulation on the ARCTIC dataset~\cite{arctic}.} SR denotes success rate. Pos., Rot., and Joint err. denote the position, rotation, and joint angle errors of the object. Contact err. denotes the average distance from the robot hand to the estimated contact points.}
\label{tab:policy_eval}
\footnotesize\setlength{\tabcolsep}{8pt}
\renewcommand{\arraystretch}{0.92}
\setlength{\aboverulesep}{0.25ex}\setlength{\belowrulesep}{0.35ex}
\begin{tabular}{l l c c c c c}
\toprule
Method & Object/Traj ID & SR (\%) $\uparrow$ & Pos. err. (cm) $\downarrow$ & Rot. err. ($^\circ$) $\downarrow$ & Joint err. ($^\circ$) $\downarrow$ & Contact err. (cm) $\downarrow$ \\
\midrule
ObjDex~\cite{objdex} & \multirow{3}{*}{Box/s04\_01} & 50.1\std{0.2} & \textbf{1.01\std{0.05}} & \textbf{1.58\std{0.24}} & 6.14\std{0.21} & -- \\
PatternDex (w/o Contact) & & 87.4\std{6.4} & 1.73\std{0.19} & 2.21\std{0.35} & 1.99\std{0.26} & -- \\
PatternDex (full) & & \textbf{99.9\std{0.1}} & 2.05\std{0.24} & 3.02\std{0.32} & \textbf{1.92\std{0.27}} & 2.35\std{0.06} \\
\midrule
ObjDex~\cite{objdex} & \multirow{3}{*}{Espresso/s01\_01} & 45.1\std{14.6} & 2.16\std{0.28} & \textbf{4.13\std{0.92}} & 2.08\std{0.26} & -- \\
PatternDex (w/o Contact) & & 47.8\std{15.0} & 2.87\std{0.30} & 6.74\std{0.95} & 9.87\std{4.03} & -- \\
PatternDex (full) & & \textbf{98.7\std{1.3}} & \textbf{1.99\std{0.17}} & 4.40\std{0.49} & \textbf{1.94\std{0.39}} & 3.61\std{0.05} \\
\midrule
ObjDex~\cite{objdex} & \multirow{3}{*}{Laptop/s09\_02} & 77.4\std{0.0} & \textbf{0.52\std{0.03}} & \textbf{1.14\std{0.08}} & 5.92\std{0.00} & -- \\
PatternDex (w/o Contact) & & 51.3\std{0.3} & 2.93\std{0.26} & 1.72\std{0.42} & 3.23\std{0.09} & -- \\
PatternDex (full) & & \textbf{94.2\std{5.1}} & 2.24\std{0.11} & 3.13\std{0.33} & \textbf{2.19\std{0.16}} & 4.24\std{0.06} \\
\midrule
ObjDex~\cite{objdex} & \multirow{3}{*}{Microwave/s04\_01} & 98.4\std{2.3} & 3.13\std{0.13} & \textbf{1.71\std{0.26}} & 2.58\std{0.32} & -- \\
PatternDex (w/o Contact) & & 76.2\std{2.7} & 3.28\std{0.18} & 2.56\std{0.39} & 2.98\std{0.31} & -- \\
PatternDex (full) & & \textbf{98.6\std{2.8}} & \textbf{2.42\std{0.16}} & 2.40\std{0.40} & \textbf{1.82\std{0.26}} & 1.92\std{0.01} \\
\midrule
ObjDex~\cite{objdex} & \multirow{3}{*}{Mixer/s08\_01} & 34.4\std{1.2} & \textbf{1.76\std{0.14}} & \textbf{3.50\std{0.28}} & 3.36\std{0.42} & -- \\
PatternDex (w/o Contact) & & 73.2\std{22.6} & 2.25\std{0.30} & 4.19\std{0.71} & 5.16\std{0.83} & -- \\
PatternDex (full) & & \textbf{99.5\std{0.7}} & 2.77\std{0.34} & 3.91\std{1.23} & \textbf{2.41\std{0.66}} & 2.91\std{0.08} \\
\midrule
ObjDex~\cite{objdex} & \multirow{3}{*}{Notebook/s08\_03} & 7.8\std{0.0} & \textbf{0.70\std{0.00}} & \textbf{2.08\std{0.01}} & 5.79\std{0.00} & -- \\
PatternDex (w/o Contact) & & 12.5\std{0.7} & 2.27\std{0.24} & 4.79\std{0.44} & \textbf{3.25\std{0.31}} & -- \\
PatternDex (full) & & \textbf{66.0\std{2.2}} & 2.94\std{0.41} & 4.68\std{0.20} & 4.06\std{0.29} & 6.65\std{0.84} \\
\midrule
ObjDex~\cite{objdex} & \multirow{3}{*}{Mean} & 52.2 & \textbf{1.55} & \textbf{2.36} & 4.31 & -- \\
PatternDex (w/o Contact) & & 58.1 & 2.55 & 3.70 & 4.41 & -- \\
PatternDex (full) & & \textbf{92.8} & 2.40 & 3.59 & \textbf{2.39} & 3.61 \\
\bottomrule
\end{tabular}
\end{table*}

%% file: sections/05_experiment/table3.tex
\begin{table*}[t]
\centering
\caption{\textbf{Cross-embodiment success rates (\%) of the guided policy.} The pattern encoder $\mathcal{E}$ is trained with Allegro and kept frozen, and only the decoder $\mathcal{D}$ and the embodiment encoder $\phi$ are fine-tuned for Shadow, Paxini, and XHand.}
\label{tab:embodiment_eval}
\footnotesize\setlength{\tabcolsep}{6pt}
\begin{tabular}{l c c c c c c c}
\toprule
Robot & Box/s04\_01 & Espresso/s01\_01 & Laptop/s09\_02 & Microwave/s04\_01 & Mixer/s08\_01 & Notebook/s08\_03 & Mean \\
\midrule
Allegro & \textbf{99.9\std{0.1}} & \textbf{98.7\std{1.3}} & \textbf{94.2\std{5.1}} & 98.6\std{2.8} & \textbf{99.5\std{0.7}} & 66.0\std{2.2} & \textbf{92.8\std{12.2}} \\
Shadow & 96.3\std{4.5} & 88.5\std{5.6} & 51.6\std{2.3} & 78.0\std{3.2} & 97.0\std{2.0} & \textbf{92.3\std{4.7}} & 84.0\std{15.8} \\
Paxini & 96.7\std{3.2} & 74.9\std{2.9} & 93.8\std{5.8} & \textbf{99.0\std{1.3}} & 12.6\std{0.0} & 64.0\std{1.7} & 73.5\std{30.0} \\
XHand & 95.4\std{5.0} & 73.8\std{3.7} & 40.0\std{6.7} & 91.9\std{7.0} & 91.0\std{3.9} & 37.3\std{1.8} & 71.6\std{24.3} \\
\bottomrule
\end{tabular}
\end{table*}

%% file: sections/figure/real_policy.tex
\begin{figure*}[t]
  \centering
  \includegraphics[width=\textwidth]{sections/figure/img/real_policy2.pdf}
  \caption{\textbf{Sim-to-Real Experiment.} (a) The wrist motions and contact points are estimated from a human demonstration video, and the guided policy is trained in the simulator with these estimates. The learned policy is then transferred to the real robot. (b) The real Allegro hand opens the microwave at steps 1, 50, and 100. The camera view and the front view are shown as insets in each snapshot.}
  \label{fig:real_policy}
  \vspace{-8pt}
\end{figure*}

%% file: sections/06_conclusion/conclusion.tex
\section{CONCLUSION}
We developed PatternDex, a model that learns the interaction pattern, a representation of the relationship between the object trajectory and the wrist motions in an embodiment-agnostic way.
We showed that the interaction pattern can be used to guide the training of a reinforcement learning policy and allow transfer to new robot hands with minimal model fine-tuning.
Since PatternDex places the robot wrists to fit the structure of each hand instead of placing them at the human wrist pose, this raises the success rate on the Allegro hand from 52.2\% to 58.1\%. 
The contact points further guide the policy on where to touch the object, which raises the mean success rate to 92.8\%, so PatternDex outperforms the baseline on all objects.
The interaction pattern learned with the Allegro hand also transfers to three other robot hands, which reach success rates above 70\%.
Moreover, a single human demonstration video is enough to train a policy for a new task and deploy it successfully on a real robot.